\pdfoutput=1

\documentclass[11pt]{article}

\usepackage{acl}
\usepackage{times}
\usepackage{latexsym}

\usepackage[T1]{fontenc}

\usepackage[utf8]{inputenc}

\usepackage{microtype}

\usepackage{inconsolata}
\usepackage{graphicx}
\usepackage{float}
\usepackage{amsmath}
\usepackage[export]{adjustbox} 
\usepackage{booktabs} 

\title{A Relation-Interactive Approach for Message Passing in Hyper-relational Knowledge Graphs}

\author{ {\bf Yonglin Jing\textsuperscript{\rm 1}} \\
	\textsuperscript{1}Qtrade AI lab \\ 
	\texttt{yj1019@nyu.edu}}

\begin{document}
	\maketitle
	\begin{abstract}
		Hyper-relational knowledge graphs (KGs) contain additional key-value pairs, providing more information about the relations. In many scenarios, the same relation can have distinct key-value pairs, making the original triple fact more recognizable and specific. Prior studies on hyper-relational KGs have established a solid standard method for hyper-relational graph encoding. In this work, we propose a message-passing-based graph encoder with global relation structure awareness ability, which we call ReSaE. Compared to the prior state-of-the-art approach, ReSaE emphasizes the interaction of relations during message passing process and optimizes the readout structure for link prediction tasks. Overall, ReSaE gives a encoding solution for hyper-relational KGs and ensures stronger performance on downstream link prediction tasks. Our experiments demonstrate that ReSaE achieves state-of-the-art performance on multiple link prediction benchmarks. Furthermore, we also analyze the influence of different model structures on model performance.
	\end{abstract}
	
	\section{Introduction}
\begin{figure}[t]
		\centering
		\includegraphics[width=0.5\textwidth]{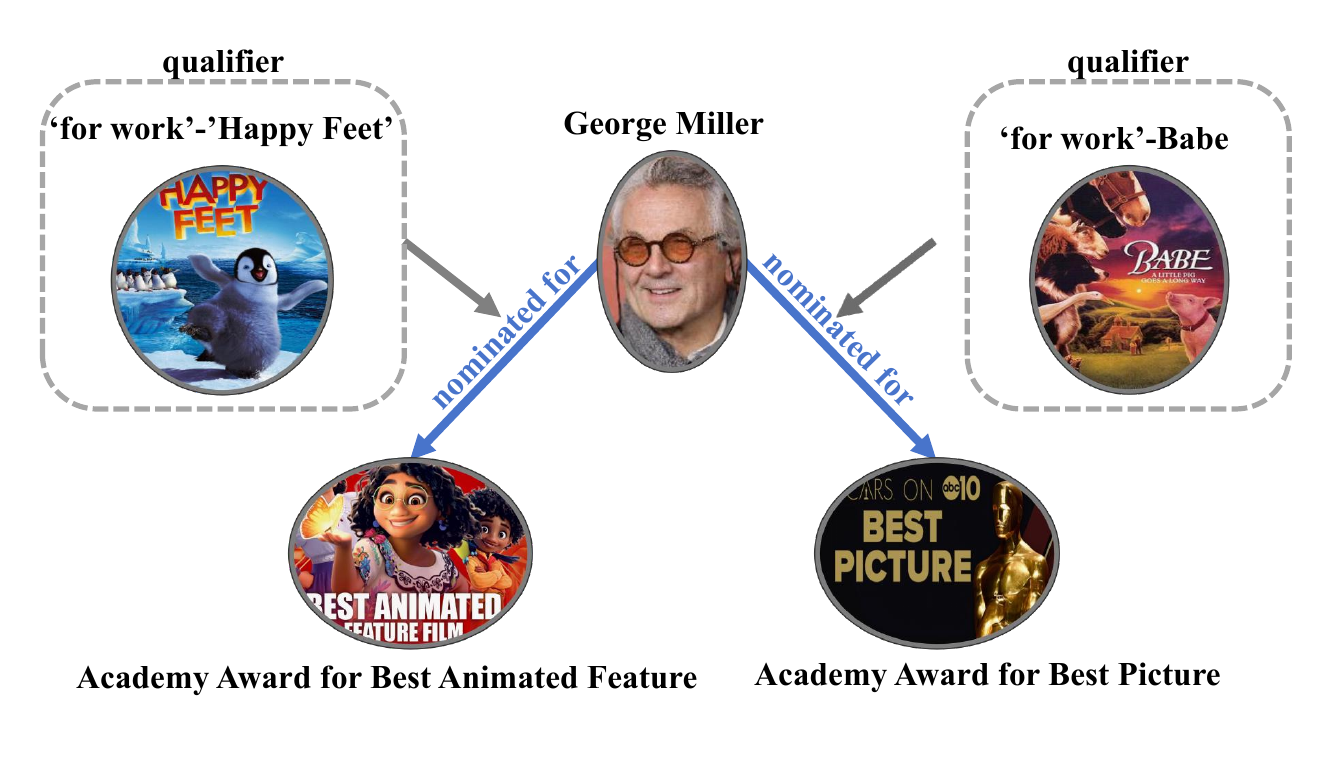}
		\caption{Two examples on hyper-relational facts. The left fact is \textit{(George Miller , nominated for, Academy Award for Best Animated Feature)} with qualifier \textit{(for work, Happy Feet)}; The right fact is \textit{(George Miller, nominated for, Academy Award for Best Pictures)} with qualifier \textit{(for work, Babe)}.} 
		\label{fig1}
	\end{figure}

	Hyper-relational knowledge graphs (KGs) differ from traditional KG settings, featuring optional information on the relation aspect. The additional information data structures can exhibit various forms, such as 
 text descriptions or web page links on the relation element. In such settings, researchers may have to consider large-scale models, \textit{e.g.} large language models (LLMs) to fully leverage the rich textual information. Another simpler and more manageable hyper-relational variant involves enriched relations with key-value pairs constraints, also known as qualifiers or triple metadata~\citep{DBLP:journals/cacm/VrandecicK14,pellissieryago,DBLP:journals/semweb/IsmayilovKALH18}. With qualifiers, we can easily distinguish hyper-relational facts. As shown in Figure~\ref{fig1}, while traditional link prediction approaches may fail to distinguish triple facts such as \textit{`George Miller'—`nominated for'—`Academy Award for Best Animated Feature'} and \textit{`George Miller'—`nominated for'—`Academy Award for Best Picture'}, these two facts can be easily distinguished by qualifiers decorating, \textit{e.g.} (\textit{`for work'—`Happy Feet'} and \textit{`for work'—`Babe'}) on relation \textit{`nominated for'}. 
	
	Qualifiers, as key-value pairs, resemble other node and edge elements (entities and relations) in knowledge graphs. Hyper-relations can accommodate an arbitrary number of pairs of qualifiers. In message passing frameworks, it is crucial to capture qualifier information in conjunction with triple fact information. While previous works mainly focus on single fact information encoding before aggregation, our method utilizes mutual information within the relation set and implements attention mechanisms during message passing. Meanwhile, we propose a neat decoder module for link prediction tasks, maintaining its permutation invariant character while enhancing performance. We validate our method through experiments on numerous link prediction benchmarks. Our code is publicly available at here\footnote{\url{https://github.com/jingyonglin/ReSaE-main}}.
 

	Our main contributions are as follows:
\begin{itemize}
    \item We propose a novel hyper-relational KG encoding architecture that can leverage the mutual information of hyper-relations during message passing.

    \item We develop a neat and efficient decoding strategy for the hyper-relational link prediction task.

    \item We investigate the influence of different attention mechanisms in the message passing process of hyper-relational KGs.
    
    \item Regarding GNN-based methods, we attain state-of-the-art performance in various hyper-relational KG link prediction benchmarks.

\end{itemize}	
	
	
	
	\begin{figure*}[htpb]
		\centering
		\includegraphics[width=\textwidth]{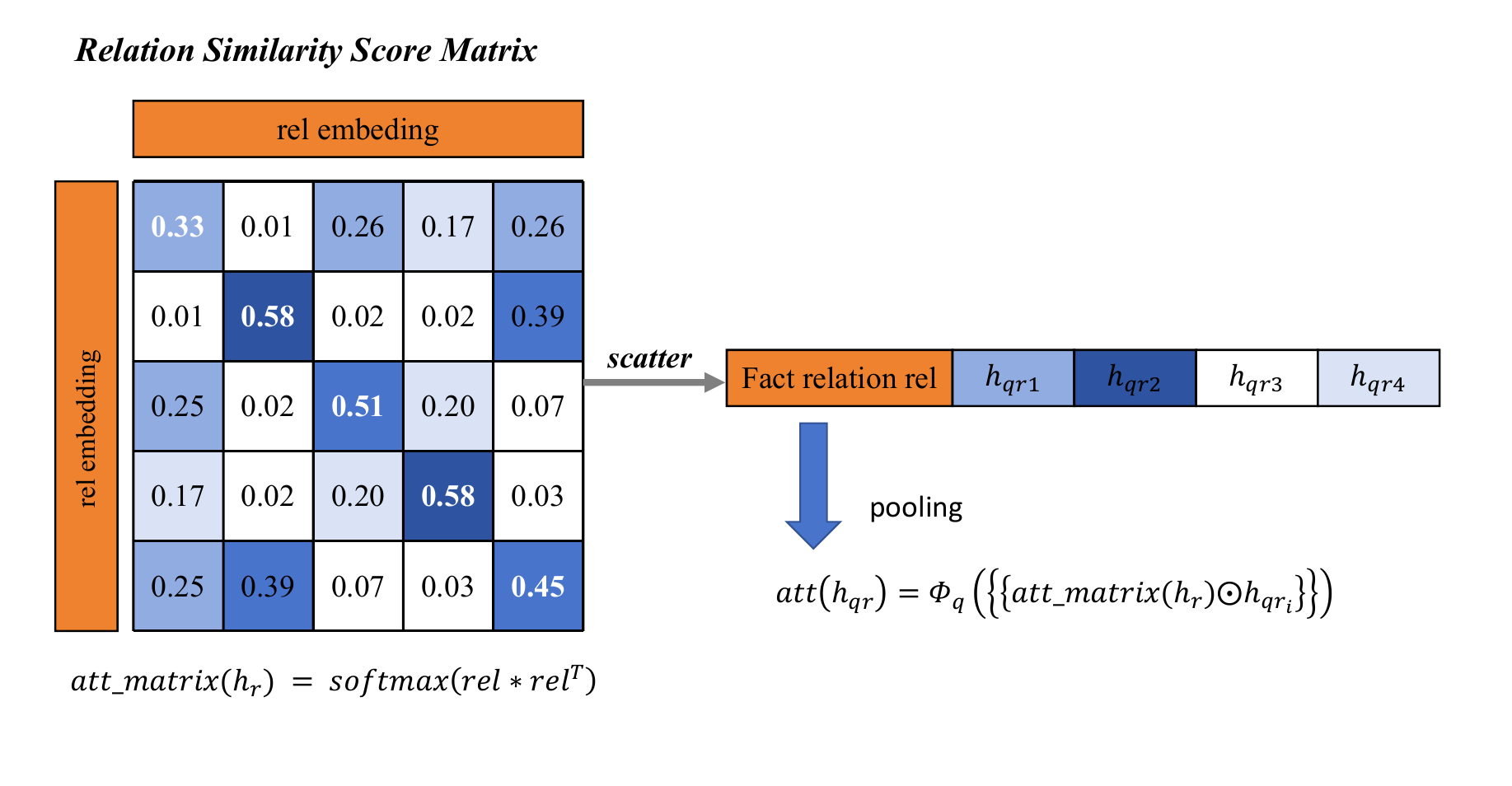}
		\caption{The self-attention illustration: On the left, we have the similarity score matrix for all relation set in a hyper-relational KG. Then for each fact, the qualifier relation weights are scattered regarding the main fact relation.} 
		\label{fig2}
	\end{figure*}
	\section{Related work}
	
	Earlier methods for hyper-relational KGs learning typically regarded qualifier information as additional information alongside the main triples. For instance, m-TransH~\citep{DBLP:conf/ijcai/WenLMCZ16} projects qualifier entity information into the main triple relation embedding space. Later approaches, such as RAE~\citep{DBLP:conf/www/ZhangLMM18} and NALP~\citep{DBLP:conf/www/GuanJWC19}, treat hyper-relational information as abstract relations or parallel fact information. However, these methods have not fully captured the characteristics of qualifier structure.
	
	HINGE~\citep{10.1145/3366423.3380257} adopts an iteratively convolved manner of composing qualifiers with main triple facts. Although it retains the hyper-relational nature of facts, HINGE operates on a triple-quintuple level that lacks the granularity of representing a certain relation instance with its qualifiers. Additionally, HINGE has to be trained sequentially in a curriculum learning~\citep{10.1145/1553374.1553380}  fashion, which requires sorting all facts in a KG in ascending order of the number of qualifiers per fact that might be prohibitively expensive for large-scale graphs.
	StarE~\citep{StarE} proposes a graph representations learning framework that can encode hyper-relational KGs with an arbitrary number of qualifiers while keeping the semantic roles of qualifiers and triples intact. It has been demonstrated to be sufficiently valid for numerous benchmarks and features a very clear model structure. However, we discovered that it faces issues such as fixed weight settings and plain pooling strategy, which may undermine the qualifier information during message passing. Notwithstanding, StarE overlooks the correlation/co-occurrence of relations in qualifiers and triples, which could play a significant role in KG representation learning.
	
	Hyper2~\citep{Hyper2} implements triple elements initialization in the Poincaré ball vectors, improving parameter efficiency and maintaining low complexity. However, it neglects the graphical structure. Other works emphasize the link prediction task and take semantic information into account, such as GRAN~\citep{GRAN} and HAHE~\citep{Luo_2023}, which enhance link prediction performance on various datasets by leveraging semantic information. However, the semantically enriched network structure mainly works on link prediction-oriented sequence representations and makes less contribution to the graph embedding part. Other knowledge graphs may not contain meaningful or faithful semantic information. Therefore, leaving semantic information aside, we focus more on GNN-based approaches for ordinary hyper-relational KGs.
	
	As for the decoder readout solution, NN4G~\citep{4773279} proposes a simple yet parameter-efficient readout method for graph classification tasks. although the method alone suffers from difficulties such as information communication between different types of nodes, the method structure presents in a clear and strictly permutation-invariant way that inspires our work.
	
	\section{Preliminaries}
	Message passing is a framework for propagating information between nodes (entities) in a graph by passing messages along the edges (relations) that connect them. The goal is to learn node representations that capture the underlying structure and relationships present in the graph. These node representations are learned iteratively by aggregating the information from the neighbors. In the case of a multi-relational graph $G=(V,R,E)$, $V$ represents the set of nodes, $R$ represents the set of relations, and $E$ denotes the set of directed edges $(s,r,o)$, where $s$, $r$ and $o$ represent subject, relation, and object, respectively. In addition, $s, o \in V$, $r \in E$, $r$ connects $s$ and $o$. Previous GNN~\citep{DBLP:conf/emnlp/MarcheggianiT17,DBLP:conf/esws/SchlichtkrullKB18} formulates the message passing process as the following:
	$$h_v^{(k)}=f(\sum \limits_{(u,r)\in N(v)}W_rh_u^{k-1}),$$
	where $f(\cdot)$ denotes non-linearity activation function, $W_r$ denotes relation specific weights, and $h_v^{k}$ denotes hidden states of node $v$ at layer $k$. The nodes $v$ and $u$ are connected with relation $r$. 
	A more parameter-efficient model CompGCN~\citep{Vashishth2020Composition-based} learns the edge vector with:
	$$h_v^{k} = f(\sum\limits_{(u,r)\in N(v)}{W_{\lambda (r)}^{(k)}\phi (h_u^{k-1},h_v^{k-1})}),$$
	where $\phi(\cdot)$ is the composition function of node and relation vector. $W_{\lambda (r)}^{(k)}$ is the parameter for aggregating the vectors of different directional types of relations (forward, inverse, loop) at layer $k$. 
	Hyper-Relational Graphs are knowledge graphs with fact tuples such as $G = (s, r, o, Q)$, where $s,r,o$ refer to subject, relation, and object respectively. $Q$ is the set of qualifier pairs, expressed as ${(qr_i, qv_i)}$, with qualifier relations $qr_i \in R$ and qualifier values $qv_i \in V$. All together $(s,r,o)$ represent the main triple,  similar to a multi-relational graph. Under hyper-relational KG settings, The number of qualifier pairs is arbitrary. For example, a triple fact: \textit{(George Miller, nominated for, Academy Award for Best Animated Feature)}, the relation part has qualifier set: \{\textit{(statement is subject of, 79th Academy Awards), (for work, Happy Feet)}\}, that present more detailed information for the main triple relation \textit{`nominated for'}.
	While performing the message passing process on hyper-relational KGs, information in qualifiers ${(qr_i, qv_i)}$ should be encoded along with the main triple. With extra information from qualifiers, recent work StarE~\citep{StarE} features the message passing update formulation as follows:
	$$h_v=f(\sum\limits_{(u,r)\in N(v)}{W_{\lambda(r)}\phi_{r}(h_u,\gamma(h_r,h_q)_{vu})}),$$	
	where $\gamma(*)$ is a function that combines the main relation representation with the representation of its qualifiers, \textit{e.g.} weighted sum, concatenation.
	
	\begin{table*}[!htbp]
		\centering
		\begin{tabular}{ccccccccccc}
			\toprule 
			~& Method &\multicolumn{3}{c}{WikiData} & \multicolumn{3}{c}{JF17K}& \multicolumn{3}{c}{WD50K} \\ 
			Metrics &  & MRR & H@1 & H@10 & MRR & H@1 & H@10 & MRR & H@1 & H@10 \\ \hline
			~ & m-TransH & 0.063 & 0.063 & 0.3 & 0.206 & 0.206 & 0.463 & - & - & - \\ \hline
			~ & RAE & 0.059 & 0.059 & 0.306 & 0.215 & 0.215 & 0.469 & - & - & - \\ \hline
			~ & NaLP-Fix & 0.42 & 0.343 & 0.556 & 0.245 & 0.185 & 0.358 & 0.177 & 0.131 & 0.264 \\ \hline
			~ & HINGE & 0.476 & 0.415 & 0.585 & 0.449 & 0.361 & 0.624 & 0.243 & 0.176 & 0.377 \\ \hline
			~ & StarE & 0.491 & 0.398 & 0.648 & 0.574 & 0.496 & 0.725 & 0.349 & 0.271 & 0.496 \\ \hline
			~ & Hyper2 & 0.461 & 0.391 & 0.597 & 0.583 & 0.5 & \textbf{ 0.746} & - & - & - \\ \hline
			~ & Hyper-transformer & 0.501 & \textbf{0.426} & 0.634 & 0.582 & 0.501 & 0.742 & 0.356 & 0.281 & 0.498 \\ \hline
			~ & ReSaE(our model)& \textbf{0.505} & 0.425 & \textbf{0.653} & \textbf{0.586} & \textbf{0.503} & 0.744 & \textbf{0.359} & \textbf{0.283} & \textbf{0.508} \\ \hline
		\end{tabular}
		\caption{\label{table1}
			Comparison of ReSaE with previous studies. The outcomes of other studies are extracted from their original papers. The optimal results are in \textbf{bold}.
		}
	\end{table*}
	
	\section{Method}
	In this section, we illustrate the general architecture of ReSaE encoder for hyper-relational Knowledge Graph representation learning and the structure used for Link Prediction. 
	All the following discussion is about single-layer network structure. In order to capture the global hyper-relational mutual information, we introduce self-attention for global relation representation. We first get the relation attention matrix by performing attention calculation on the whole relation set ($R$ as mentioned in Preliminaries) as follows:
   \begin{equation}
	\begin{aligned}
att\_matrix(h_{r})= softmax(\frac{h_rh_r^T}{\sqrt{d}}),
	\end{aligned}
 \label{eq:1}
 \end{equation}
	where $h_r$ is the relation representation and $d$ denotes the embedding dimension. Here we do not perform a linear projection before the inner product, with detailed reasons presented in section \ref{section.5}. For more analysis of attention mechanism and other possible attention variants, which we will also discuss in section \ref{section.5}. 
	
	Then on the message passing stage, we cast the attention weight on the qualifier relation part according to their main triple relation $h_r$. And then pooling the qualifier relation features for each fact, \textit{e.g.} sum and mean. Thus, we obtain the attention-weighted qualifier relation representation for each hyper-relational fact, as shown in the formula below:
  \begin{equation}
	\begin{aligned}
att(h_{qr})=\Phi_q( att\_matrix(h_{r})\odot h_{qr}),
	\end{aligned}
 \label{eq:2}
 \end{equation}	
	where $\Phi_q$ denotes the pooling method and $\odot$ is the casting process.
	The attention calculation is illustrated in Figure \ref{fig2}.
	
	Along with attention output, we also take simple pooling over qualifier relations and entities, so the features for message passing would be:
  \begin{equation}
	\begin{aligned}
h_{hyper}=[h_u,h_r,\gamma_r(h_{qr}),\gamma_v(h_{qv}),att(h_{qr})],
	\end{aligned}
 \label{eq:3}
 \end{equation}
	where $h_{hyper}$ denotes the hyper-relation fact representation. $\gamma_r(\cdot)$ and $\gamma_v(\cdot)$ denote pooling over qualifier relation and qualifier entity, respectively. 
	Optionally, qualifier relation information can also be seen as pooling over $\gamma_r(h_{qr}),att(h_{qr})$. Under this setting, equation \ref{eq:3} can be rewritten as:
  \begin{equation}
	\begin{aligned}
h_{hyper}=[h_u,h_r,\gamma_v(h_{qv}),\Phi_{q2}(\gamma_r(h_{qr}),att(h_{qr}))],
	\end{aligned}
 \label{eq:4}
\end{equation}
	where $\Phi_{q2}$ denotes the pooling method.
	Then the message passing aggregation function can be written as equation \ref{eq:5}:
 \begin{equation}
	\begin{aligned}
		h_u &=  W_{\lambda (r)}\Phi_{r} (h_{hyper}) \\
  h_v &= f(\alpha \cdot \sum\limits_{(u,r) \in N(v)}{h_u}+\beta h_v), 
	\end{aligned}
 \label{eq:5}
\end{equation}
	where $\Phi_r(\cdot)$ is a function that aggregates hyper-relational fact information concerning the relation type $r$, which can be concatenation, average, or weighted sum. Then the outcome is transformed by the weight $W_\lambda$ to obtain hyper-relational fact representations corresponding to its directional type $\lambda$ (forward, inverse, loop). 
	
\begin{figure}[htpb]
		\centering
		\includegraphics[width=0.5\textwidth]{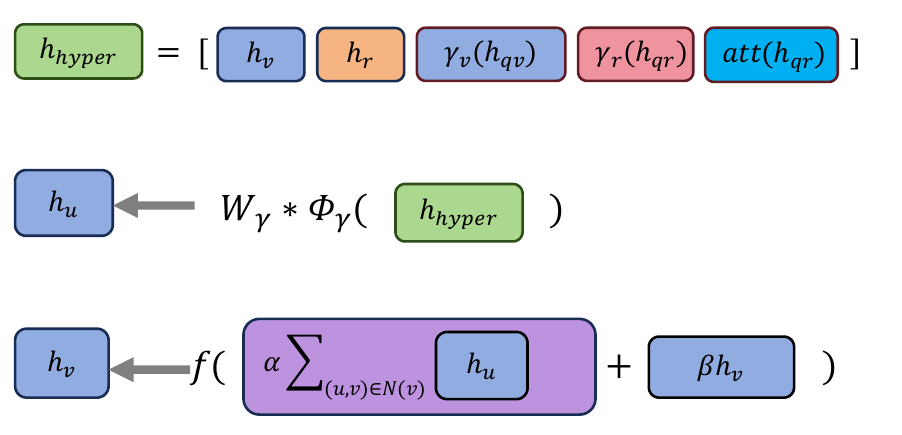}
		\caption{the hyper-relational fact entity update process. $h_{hyper}$ contains information of fact entity, main relation, qualifier entities aggregation, qualifier relations aggregation, aggregation of $att(h_{qr})$. $h_v^\gamma$ is linear projection of  $h_{hyper}$ aggregation output and $f$ denotes the activation function. } 
		\label{fig3}
	\end{figure}

	Finally, the node (entity) representations are obtained by taking the weighted sum of itself and the sum of its neighbor representations. $f(\cdot)$ denotes the non-linearity activate function, $\alpha$ and $\beta$ are trainable parameters. 
	The overview of the node updating process is illustrated in Figure \ref{fig3}.

	During the relation update stage, to better capture the global relation distribution information, we introduce a co-occurrence information matrix for the relation set.
	The matrix calculation is through the progress below:
	
	In case of single relation type $\gamma$, we sum up all co-appear time of relation $(r_i, r_j)$. (\textit{e.g.} the main fact relation is $r_i$, if $r_j$ in its qualifier, the count plus 1) We note the summation as $N_{coo}^\gamma(r_i^\gamma,r_j^\gamma)$. And for each relation $r_i$, we sum up $N_{coo}^\gamma(r_i^\gamma,r_j^\gamma)$ along $j$, note as $N_{coo}r_{i}^\gamma$. Then normalize $N_{coo}^\gamma(r_i,r_j)$ by $N_{coo}r_{i}^\gamma$, we get the co-occurrence information matrix for relations Note as $coo_{r_{i,j}}^\gamma$, or short as $coo_r^\gamma$.
	The whole calculation function is as below:
	$$ coo_{r_{i,j}}^\gamma = \frac{N_{coo}(r_i^\gamma,r_j^\gamma) }{mean(N_{coo}r_{i}^\gamma)} $$
	
	Thus, the relation of type $\gamma$ is updated through the following function:
	
	$$h_r^\gamma = Avg(\psi_r(h_r^\gamma,h_r^\gamma, coo_r^\gamma))$$
	$$\psi_r(h_r^\gamma,h_r^\gamma, coo_r^\gamma) =\alpha h_r^\gamma+\beta W(h_r^\gamma \cdot coo_r^\gamma),$$
	where $W(h_r^\gamma \cdot coo_r^\gamma)$ is linear transformation over multiplication of $h_r^\gamma$ and co-occurrence information of $h_r^\gamma$. Avg() stands for mean function, $\psi_r$ is the weight sum function between $h_r^\gamma$ and $W(h_r^\gamma\cdot coo_r^\gamma)$.

\begin{figure}[htpb]
		\centering
		\includegraphics[width=0.5\textwidth]{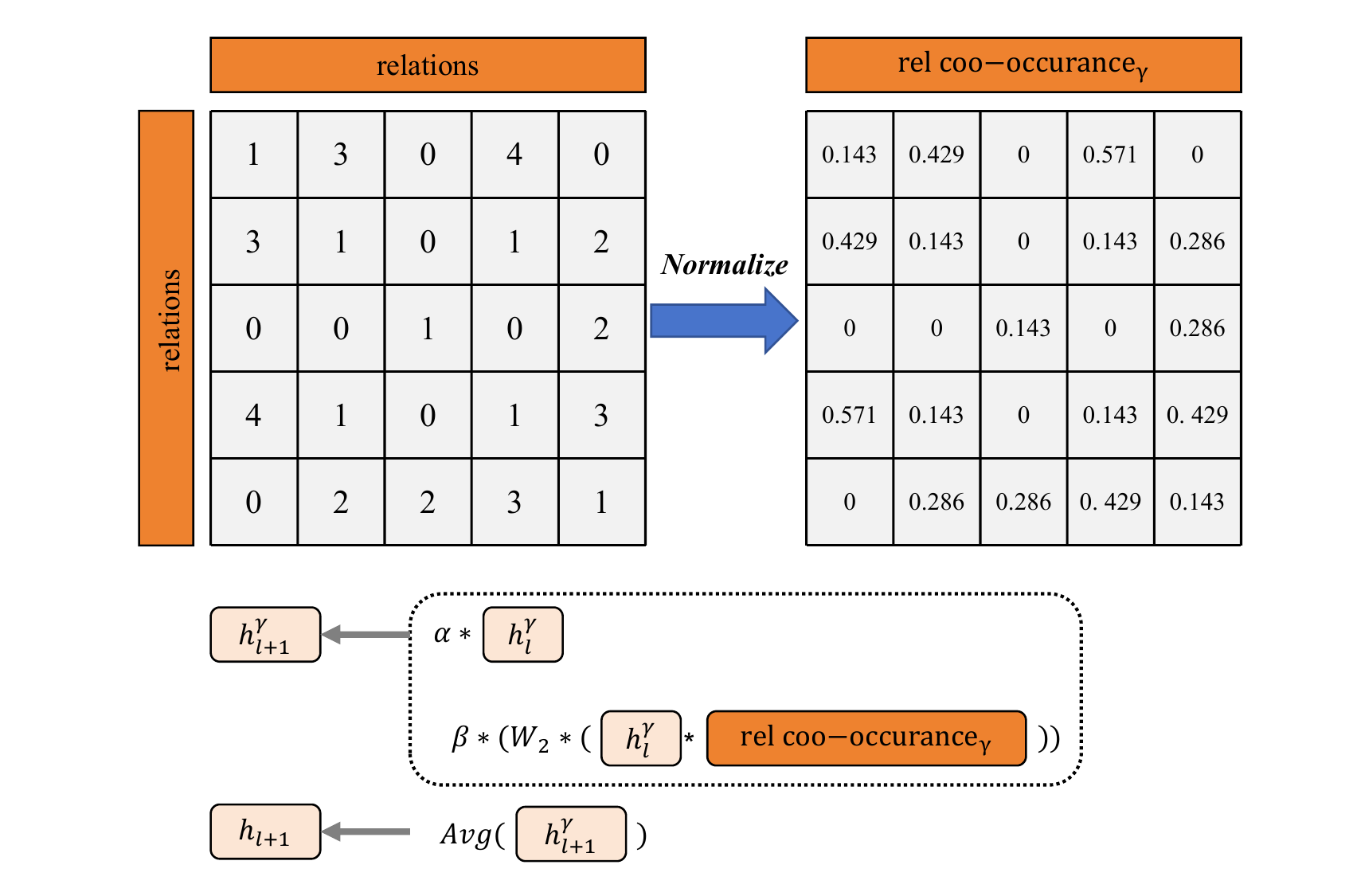}
		\caption{The relation update process: On the top left, each cell represents the co-occurrence time of two relations in all hyper-relational facts. We then normalize each cell by dividing it by mean of row sum, resulting in the matrix on the top right. Each type (relation direction) of relations is updated by weight sum of itself and linear projection of (relation * coo-occurrence matrix).  Finally, the relation representation is updated by taking the average of its type.} 
		\label{fig4}
	\end{figure}

	The representation of the relation set is updated by averaging over different directional type relations, followed by activation:
	$$h_r = act_r(avg(h_r^\gamma)),$$
	where $act_r(\cdot)$ is the activation function (linear or non-linear activate function).
	The relation update process of a ReSaE layer is illustrated in Figure \ref{fig4}. 
	
	\begin{table*}[!htbp]
		\centering
		\resizebox{\linewidth}{!}{
			\begin{tabular}{cccccccccccccc}
				\toprule 
				~& DataSet &\multicolumn{3}{c}{WD50k} & \multicolumn{3}{c}{WD50K\_33}& \multicolumn{3}{c}{WD50K\_66} & \multicolumn{3}{c}{WD50K\_100}\\ 
				Metrics &  & MRR & H@1 & H@10 & MRR & H@1 & H@10 & MRR & H@1 & H@10 & MRR & H@1 & H@10\\ \hline
				~ & HINGE & 0.243 &	0.176 &	0.377&	0.253&	0.19&	0.372&	0.378&	0.307&	0.512&	0.492	&0.417&	0.636\\ \hline
				~ & Transformer(H) & 0.286&	0.222&	0.406&	0.276&	0.227&	0.371&	0.404&	0.352&	0.502&	0.562&	0.499&	0.677\\ \hline
				~ & StarE & 0.349&	0.271&	0.496&	0.331&	0.268&	0.451&	0.481&	0.42&	0.594&	0.654&	0.588&	0.777\\ \hline
				~ & ReSaE& \textbf{0.358}&	\textbf{0.28}&\textbf{0.507}&	\textbf{0.336}&	\textbf{0.273}&	\textbf{0.453}&	\textbf{0.493}&	\textbf{0.433}&	\textbf{0.602}&	\textbf{0.668}&	\textbf{0.605}&	\textbf{0.785} \\ \hline
		\end{tabular}}
		\caption{\label{table2}
			Link prediction on WD50K graphs with different ratio of qualifiers. The best results are in \textbf{bold}
		}
	\end{table*}

\begin{figure}[htpb]
		\centering
		\includegraphics[width=0.5\textwidth]{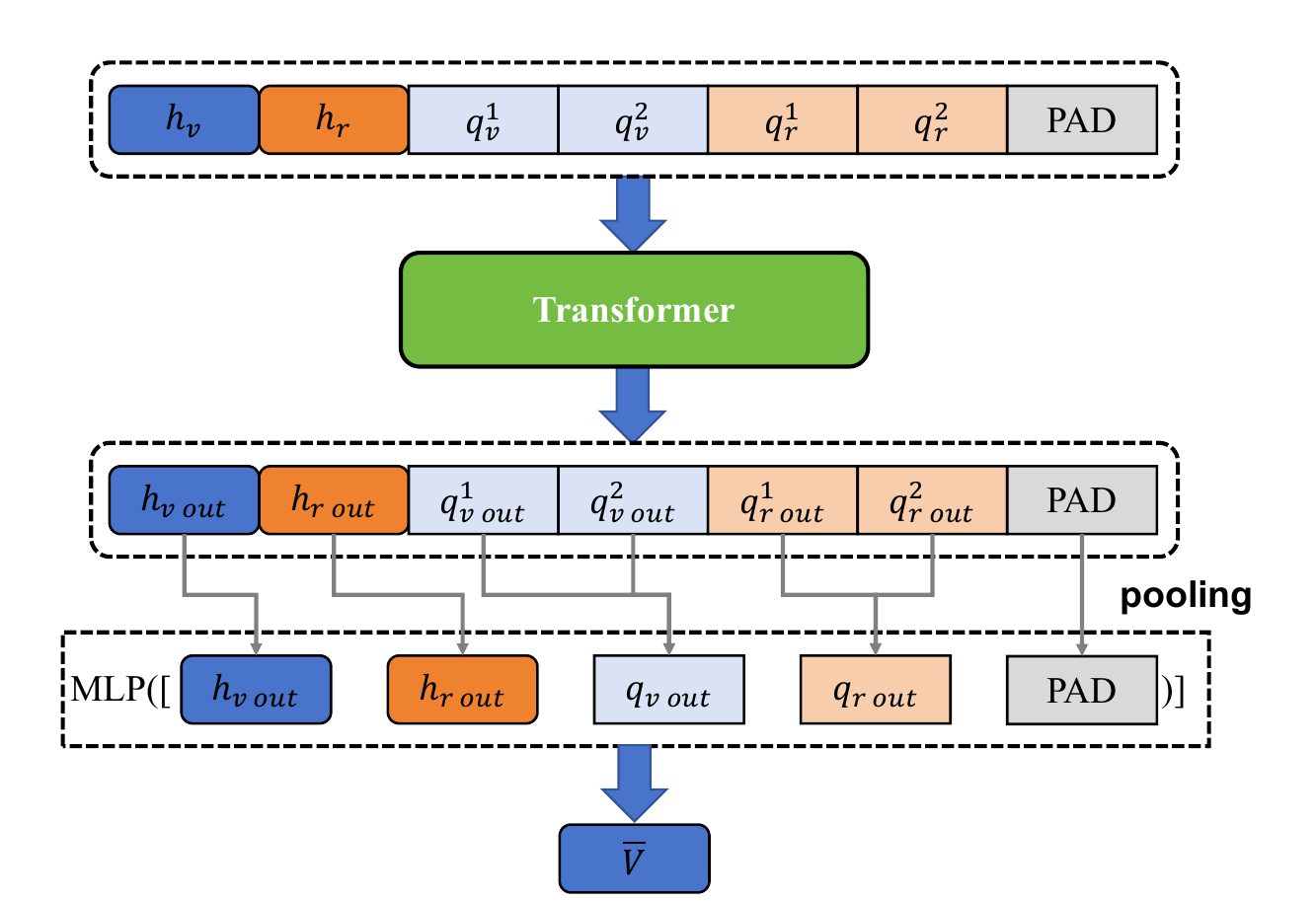}
		\caption{The decoder architecture. Hyper-relational fact statements go through transformer layers. Subsequently, the output sequence is pooled based on their type. The final hidden state is attained through a fully connected layer.} 
		\label{fig5}
	\end{figure}

	For link prediction task, we take similar decoder setting with StarE~\citep{StarE}, which is a transformer~\citep{DBLP:conf/nips/VaswaniSPUJGKP17} based decoder. We re-arrange the input and readout module. As for downstream tasks, we still want to preserve the graph learning characteristics, which is permutation invariance. So we remove the positional embedding input.
	Furthermore, inspired by NN4G~\citep{4773279}, we construct a new type-wise pooling readout module as:
	
	$$out = MLP([\phi_{type_i}(h_{type_i}),...]),$$
	where $\phi_{type_i}$ would be pooling (mean or max) over $i$-th type of node. There are five types of nodes in total: relation, qualifier entity, qualifier relation, and mask. So the longest possible sequence would be: $$[h_v, h_r, \phi_{q\ edge}(h_{qv}), \phi_{q\ rel}(h_{qr}), \phi_{pad}(h_{pad})],$$ where $[\cdot]$ denotes concatenation along the last dimension. Notice the mean function on $h_v$ and $h_r$ remain unchanged. 
	This readout module maintains the permutation invariant feature and simultaneously captures more information than a simple pooling(\textit{e.g.} mean, sum, max) approach. We obtain the final output by applying a Multi-Layer Perceptron (MLP) over the concatenated hidden states. We use this decoder for one-vs-all classification on the link prediction task. Additionally, we conduct experiments with other decoder/readout variants, which we will discuss in Section \ref{section.5}.
	The decoder structure is illustrated in Figure \ref{fig5}. 

	\section{Experiments and Analysis}\label{section.5}
	In this section, we evaluate the effectiveness of our approach.
	\subsection{Datasets:}
	We experiments on three hyper-relational datasets: JF17K~\citep{DBLP:conf/ijcai/WenLMCZ16}, WikiPeople~\citep{DBLP:conf/www/GuanJWC19}, and WD50K~\citep{StarE}. Among them we conduct WD50K experiments on different qualifier ratio settings, which are WD50K (33\%), WD50K (66\%), and WD50k (100\%), indicating different ratio of hyper-relational triple. Literal context statements in WikiPeople are filtered out. JF17K is extracted from Freebase~\citep{Freebase}. The main statistics are shown in Appendix Table \ref{table5}.
	
	\subsection{Baseline}
	
	We mainly compare ReSaE with GNN-based methods under similar parameter sizes, including:
	(i) m-TransH~\citep{DBLP:conf/ijcai/WenLMCZ16}, (ii) RAE~\citep{DBLP:conf/www/ZhangLMM18},(iii) NaLP~\citep{DBLP:conf/www/GuanJWC19}, (iv) NeuInfer~\citep{NeuInfer}, (v) HINGE~\citep{10.1145/3366423.3380257}, (vi) StarE~\citep{StarE} (vii) Hyper2~\citep{Hyper2}, and (viii) HyTransformer~\citep{HyTransformer}.
	\subsection{Metrics}
	For link prediction, we report Mean Reciprocal Rank (short as MRR, introduced in~\cite{DBLP:conf/nips/BordesUGWY13}), and Hits@{1, 10}. Hits@K measures the proportion of correct entities ranked in the top K. The results are averaged over test dataset and over predicting missing head and tail entities. Generally, a good model is expected to achieve higher MRR and Hits@N.
	\subsection{Ablations}
	To verify the significance of our method, we conduct experiments on some simplified model variants by replacing or removing module parts from the full model. ReSaE w/o coo, and ReSaE w/o att are methods without relation coo-matrix, and without relation self-attention module respectively. Transformer+MEAN pool refers to the method that simply performs mean pooling of the transformer output without type-wise readout mechanism. All variants are trained under the same setting as the main experiment, except for the differences in their respective modules.
	\subsection{Training Subject}
	For the task of link prediction, the model is trained through $1-N$ cross-entropy loss as function bellow:
	$$ loss = \sum\limits_{i}^{N}y_ilogP_i,$$
	where $y_i$ is the $t$-th entry of the label $y$. $P_i$ represents the sigmoid score calculated based on the one-vs-all cosine similarity between the final hidden states and all entity representations.
	
	\subsection{Training Settings}
	For ReSaE, we conduct all experiments with the same hidden size of 200, and use the Adam~\citep{DBLP:journals/corr/KingmaB14} optimizer with learning rate of $10^{-4}$. Label smoothing~\citep{DBLP:conf/aaai/DettmersMS018,Vashishth2020Composition-based} is set to 0.1. The model was trained on varying batch sizes and epochs, with detailed parameter settings presented in Table \ref{table4} in the Appendix. For related work in comparison, we reuse the metrics records of previous works in our experiment result table.
	
	\subsection{General Performance}
	We compare our methods to prior approaches on the link prediction task. Experimental results in Table \ref{table1} demonstrate that in the case of MMR, our method outperforms the previous method and achieves the state-of-the-art performance among all GNN-based methods. Nevertheless, our method only performs slightly better than HyperTransformer~\citep{HyTransformer} and underperforms on some metrics on some metrics (\textit{e.g.} H@1, H@10).
	In the case of WD50K, our method outperforms most metrics compared to several previous methods, demonstrating the hyper-relational encoding capability of our approach. 
	
	As shown in Table \ref{table2}, we achieve a greater performance improvement over WD50K with a higher ratio of hyper-relational triples, indicating that our method performs better with hyper-relational facts than traditional ones.
	
	We also acquired the experimental records of the ablated ReSaE variants as presented in Table \ref{table3}. This suggests that removing any one of the modules would result in some decrease in the final model performance. The experimental results demonstrate the necessity of all corresponding components and verify the effectiveness of the ReSaE architecture.
	
	\begin{table*}[!htbp]
		\centering
		\begin{tabular}{ccccc}
			\toprule 
			~& &\multicolumn{3}{c}{WD50K\_100} \\ 
			Encoder Method & Decoder Method & MRR & H@1 & H@10 \\ \hline
			ReSaE w/o coo & Transformer+typewise pooling &0.659	&0.597	&0.78 \\ \hline
			ReSaE with linear-projection att & Transformer+typewise pooling &0.662	&0.599 &0.782 \\ \hline
			ReSaE w/o att & Transformer+typewise pooling &0.657 &0.594 &0.771 \\ \hline
			ReSaE & GAT &0.635	&0.57	&0.721 \\ \hline
			ReSaE & Transformer+MEAN pool &0.662	&0.601	&0.779 \\ \hline
			ReSaE & Transformer+typewise pooling & \textbf{0.668}	&\textbf{0.605}	&\textbf{0.785} \\ \hline
		\end{tabular}
		\caption{\label{table3}
			ablation results of model variants. Decoder and encoder variants link prediction performance comparison. The best results are in \textbf{bold}.
		}
	\end{table*}
	
	\subsection{Analysis for qualifier aggregation}
	The qualifier set $\{(qr_i, qb_i)\}$ for a single hyper-relational typically consists of numerous pairs of (relation, entity). Message passing necessitates organizing the qualifier representation into a fixed-length hidden state. Conventional aggregation/pooling strategies such as addition, mean, max, and rotation~\citep{DBLP:conf/iclr/SunDNT19} have the following two main assumptions:
	
	1. Entity and relation share the same embedding space or co-exist in the same complex hidden space (for rotation setting).
	
	2. For most hyper-relational facts, every pair of qualifiers is supposed to be nearly as important.
	
	The above two assumptions may be true for regular Knowledge Graph (KG) scenarios; however, they might not hold for hyper-relational KGs. In hyper-relational KGs, qualifiers serve as the `decoration' of relations, and the (relation, entity) pairs are quite distinctive. Traditional pooling strategies such as \texttt{sum}, \texttt{mean}, and \texttt{max} can easily lead to ambiguity. Furthermore, qualifier entities and relation pairs usually describe information that is much more complex than directional information. These cannot be treated as complex embedding spaces as in classical KG learning approaches, \textit{e.g.} ComplE~\citep{Lacroix2020Tensor}. 
	
	Intuitively, we need a separate space to preserve qualifier information and a weighted aggregation instead of a regular aggregation approach. Under these circumstances, the attention mechanism is employed, and qualifier hidden states are aggregated separately with the main triple parts.
	
	Our intuition was validated by the ablation studies. The attention matrix scores also suggest the necessity of our weighted aggregation strategy. For instance, WD50K contains the following relations:

\begin{itemize}
    \item \textit{$r_a$ (nominated for)}
    \item \textit{$r_b$ (statement is subject of)}  
    \item \textit{$r_c$ (for work)} 
\end{itemize}

	These three relations co-occur in a single hyper-relational fact, as shown below:
	
	\textit{(George Miller , nominated for, Academy Award for Best Animated Feature)} 
	
	with qualifier:
	
	\textit{\{(statement is subject of, 79th Academy Awards), (for work, Happy Feet)\}},
	
	in which the main triple relation is $r_a$,  $r_b$ and $r_c$ appear in its qualifiers. After training loops, the attention score between $r_a$ and $r_b$ is 0.214 while the attention score between $r_a$ and $r_c$ is 0.710. This follows the designed expectation because the semantic meaning of $r_c$ is much more related to $r_a$ than $r_b$ is. Although $r_b$ is a much more frequent qualifier relation in WD50K, it does not necessarily mean it should be equally important to other qualifier relations in each case. 
	
	\subsection{Relation update different with conventional KGs}
	The representations of relation get updated after that of the nodes (entity embedding). Unlike node representations, not a large number of relations are typically involved in hyper-relational KG settings. Instead of regular linear updates, we propose updating relations by leveraging the structural information in the hyper-relational KGs. Since qualifiers carry relations that also belong to relation set $R$. Intuitively, co-occurrence data might be beneficial. In the case of WD50k\_100, the experiment shows that neglecting relation co-occurrences (in the ablation study the model variance ReSaE w/o coo) while performing relation update results in a 2\%  drop in link prediction performance. 
	
	\subsection{Analysis of attention mechanism}
	Since there is much to explore regarding the attention mechanism, we also conduct experiments on several self-attention model variants on the WD50k\_100 dataset. We experiment with linear projection and multi-head module while calculating attention score just as the classical transformer does. Nevertheless, we do not notice obvious performance differences. We argue that the most likely reason is that the size of the relation set is not as large as in other attention implementation scenarios, such as conventional NLP tasks. The largest relation set in our work is 532 (WD50K). Therefore, the embedding size (200 in our main experiment) is sufficient enough to preserve the relation information. In this case, sophisticated attention calculation may lead to over-parameterization issues. Therefore we eventually adopt the single-head self-attention module without linear projection. In future research, we may investigate under what circumstances attention variants would become necessary.
	
	\subsection{Importance of decoder and readout}
	The choice of decoder is crucial for the link prediction task. We also examine different decoder variants, such as the MLP decoder and the GAT~\citep{DBLP:conf/iclr/VelickovicCCRLB18} decoder. For the MLP decoder, we simply concatenate the representations of all elemental nodes and pass them through a fully connected layer. For the GAT decoder, we introduced a virtual readout node that was connected to all other nodes. We conduct these variant experiments on WD50k\_100 while maintaining the same training settings as our main approach. Surprisingly, according to the results shown in Table \ref{table3}, neither the GAT decoder nor the MLP decoder(MEAN pool) improved the LP performance. 
	
	Additionally, we test the transformer decoder with and without positional encoding and token type ids encoding. Model variant with positional encoding demonstrates a performance decline. We believe that the ReSaE encoder already enhances graph representations with structural and character features, making it redundant to add positional information during the decoding process. 
	
	Furthermore, we believe that for downstream tasks, permutation invariance should still be maintained. ReSaE readout maintains permutation invariance by not adding positional encoding and employing a type-wise pooling strategy, and indeed, we observe performance increases in our experiments.
	
	Throughout our experiments, we have managed to achieve a satisfactory result: the decoder section remains straightforward and elegant, while the robust encoder ensures that the hyper-relational KG representations are rich in information for downstream tasks.

	\section{Conclusion}
	In this paper, we present ReSaE, a message-passing framework for learning hyper-relational Knowledge Graphs (KGs). ReSaE leverages self-attention during message passing and co-occurrence information when updating relation representations. Additionally, ReSaE offers a straightforward yet effective decoder readout solution for hyper-relational KG link prediction tasks. The network layers are designed to be straightforward and are not constrained by data structure settings. Experimental results extensively show that, compared to existing GNN-based approaches, our method achieves state-of-the-art performance on multiple benchmarks. Furthermore, we provide detailed analysis of different module parts, explaining our intuition and presenting relevant experimental results to validate our design. 
	In future work, we intend to investigate more advanced learning techniques that can effectively capture the sparse connections between qualifier relations and entities.

	\bibliography{anthology,custom}

\begin{thebibliography}{27}
\expandafter\ifx\csname natexlab\endcsname\relax\def\natexlab#1{#1}\fi

\bibitem[{Bengio et~al.(2009)Bengio, Louradour, Collobert, and
  Weston}]{10.1145/1553374.1553380}
Yoshua Bengio, J\'{e}r\^{o}me Louradour, Ronan Collobert, and Jason Weston.
  2009.
\newblock Curriculum learning.
\newblock In \emph{Proceedings of the 26th Annual International Conference on
  Machine Learning}, page 41–48.

\bibitem[{Bollacker et~al.(2008)Bollacker, Evans, Paritosh, Sturge, and
  Taylor}]{Freebase}
Kurt Bollacker, Colin Evans, Praveen Paritosh, Tim Sturge, and Jamie Taylor.
  2008.
\newblock \href {https://doi.org/10.1145/1376616.1376746} {Freebase: A
  collaboratively created graph database for structuring human knowledge}.
\newblock In \emph{Proceedings of the 2008 ACM SIGMOD International Conference
  on Management of Data}, SIGMOD '08, page 1247–1250, New York, NY, USA.
  Association for Computing Machinery.

\bibitem[{Bordes et~al.(2013)Bordes, Usunier, Garc{\'{\i}}a{-}Dur{\'{a}}n,
  Weston, and Yakhnenko}]{DBLP:conf/nips/BordesUGWY13}
Antoine Bordes, Nicolas Usunier, Alberto Garc{\'{\i}}a{-}Dur{\'{a}}n, Jason
  Weston, and Oksana Yakhnenko. 2013.
\newblock Translating embeddings for modeling multi-relational data.
\newblock In \emph{Advances in Neural Information Processing Systems}, pages
  2787--2795.

\bibitem[{Dettmers et~al.(2018)Dettmers, Minervini, Stenetorp, and
  Riedel}]{DBLP:conf/aaai/DettmersMS018}
Tim Dettmers, Pasquale Minervini, Pontus Stenetorp, and Sebastian Riedel. 2018.
\newblock Convolutional 2d knowledge graph embeddings.
\newblock In \emph{Proceedings of the Thirty-Second {AAAI} Conference on
  Artificial Intelligence}, pages 1811--1818.

\bibitem[{Fey and Lenssen(2019)}]{DBLP:journals/corr/abs-1903-02428}
Matthias Fey and Jan~Eric Lenssen. 2019.
\newblock Fast graph representation learning with pytorch geometric.
\newblock \emph{CoRR}, abs/1903.02428.

\bibitem[{Galkin et~al.(2020)Galkin, Trivedi, Maheshwari, Usbeck, and
  Lehmann}]{StarE}
Mikhail Galkin, Priyansh Trivedi, Gaurav Maheshwari, Ricardo Usbeck, and Jens
  Lehmann. 2020.
\newblock \href {https://doi.org/10.18653/v1/2020.emnlp-main.596} {Message
  passing for hyper-relational knowledge graphs}.
\newblock In \emph{Proceedings of the 2020 Conference on Empirical Methods in
  Natural Language Processing (EMNLP)}, pages 7346--7359, Online. Association
  for Computational Linguistics.

\bibitem[{Guan et~al.(2020)Guan, Jin, Guo, Wang, and Cheng}]{NeuInfer}
Saiping Guan, Xiaolong Jin, Jiafeng Guo, Yuanzhuo Wang, and Xueqi Cheng. 2020.
\newblock \href {https://doi.org/10.18653/v1/2020.acl-main.546} {{N}eu{I}nfer:
  Knowledge inference on {N}-ary facts}.
\newblock In \emph{Proceedings of the 58th Annual Meeting of the Association
  for Computational Linguistics}, pages 6141--6151, Online. Association for
  Computational Linguistics.

\bibitem[{Guan et~al.(2019)Guan, Jin, Wang, and
  Cheng}]{DBLP:conf/www/GuanJWC19}
Saiping Guan, Xiaolong Jin, Yuanzhuo Wang, and Xueqi Cheng. 2019.
\newblock Link prediction on n-ary relational data.
\newblock In \emph{The World Wide Web Conference, {WWW} 2019}, pages 583--593.

\bibitem[{Ismayilov et~al.(2018)Ismayilov, Kontokostas, Auer, Lehmann, and
  Hellmann}]{DBLP:journals/semweb/IsmayilovKALH18}
Ali Ismayilov, Dimitris Kontokostas, S{\"{o}}ren Auer, Jens Lehmann, and
  Sebastian Hellmann. 2018.
\newblock Wikidata through the eyes of dbpedia.
\newblock \emph{Semantic Web}, 9(4):493--503.

\bibitem[{Kingma and Ba(2015)}]{DBLP:journals/corr/KingmaB14}
Diederik~P. Kingma and Jimmy Ba. 2015.
\newblock Adam: {A} method for stochastic optimization.
\newblock In \emph{3rd International Conference on Learning Representations,
  {ICLR} 2015}.

\bibitem[{Lacroix et~al.(2020)Lacroix, Obozinski, and
  Usunier}]{Lacroix2020Tensor}
Timothée Lacroix, Guillaume Obozinski, and Nicolas Usunier. 2020.
\newblock Tensor decompositions for temporal knowledge base completion.
\newblock In \emph{International Conference on Learning Representations}.

\bibitem[{Luo et~al.(2023)Luo, E, Yang, Guo, Sun, Yao, Tang, Wan, Song, and
  Lin}]{Luo_2023}
Haoran Luo, Haihong E, Yuhao Yang, Yikai Guo, Mingzhi Sun, Tianyu Yao, Zichen
  Tang, Kaiyang Wan, Meina Song, and Wei Lin. 2023.
\newblock \href {https://doi.org/10.18653/v1/2023.acl-long.450} {Hahe:
  Hierarchical attention for hyper-relational knowledge graphs in global and
  local level}.
\newblock In \emph{Proceedings of the 61st Annual Meeting of the Association
  for Computational Linguistics (Volume 1: Long Papers)}. Association for
  Computational Linguistics.

\bibitem[{Marcheggiani and Titov(2017)}]{DBLP:conf/emnlp/MarcheggianiT17}
Diego Marcheggiani and Ivan Titov. 2017.
\newblock Encoding sentences with graph convolutional networks for semantic
  role labeling.
\newblock In \emph{Proceedings of the 2017 Conference on Empirical Methods in
  Natural Language Processing, {EMNLP} 2017, Copenhagen, Denmark, September
  9-11, 2017}, pages 1506--1515.

\bibitem[{Micheli(2009)}]{4773279}
Alessio Micheli. 2009.
\newblock \href {https://doi.org/10.1109/TNN.2008.2010350} {Neural network for
  graphs: A contextual constructive approach}.
\newblock \emph{IEEE Transactions on Neural Networks}, 20(3):498--511.

\bibitem[{Pellissier-Tanon et~al.(2020)Pellissier-Tanon, Weikum, and
  Suchanek}]{pellissieryago}
Thomas Pellissier-Tanon, Gerhard Weikum, and Fabian Suchanek. 2020.
\newblock Yago 4: A reason-able knowledge base.
\newblock In \emph{Extended Semantic Web Conference, {ESWC} 2020}.

\bibitem[{Rosso et~al.(2020)Rosso, Yang, and
  Cudr\'{e}-Mauroux}]{10.1145/3366423.3380257}
Paolo Rosso, Dingqi Yang, and Philippe Cudr\'{e}-Mauroux. 2020.
\newblock Beyond triplets: Hyper-relational knowledge graph embedding for link
  prediction.
\newblock In \emph{Proceedings of The Web Conference 2020}, page 1885–1896.

\bibitem[{Schlichtkrull et~al.(2018)Schlichtkrull, Kipf, Bloem, van~den Berg,
  Titov, and Welling}]{DBLP:conf/esws/SchlichtkrullKB18}
Michael~Sejr Schlichtkrull, Thomas~N. Kipf, Peter Bloem, Rianne van~den Berg,
  Ivan Titov, and Max Welling. 2018.
\newblock Modeling relational data with graph convolutional networks.
\newblock In \emph{The Semantic Web - 15th International Conference, {ESWC}
  2018, Heraklion, Crete, Greece, June 3-7, 2018, Proceedings}, pages 593--607.

\bibitem[{Sun et~al.(2019)Sun, Deng, Nie, and Tang}]{DBLP:conf/iclr/SunDNT19}
Zhiqing Sun, Zhi{-}Hong Deng, Jian{-}Yun Nie, and Jian Tang. 2019.
\newblock Rotate: Knowledge graph embedding by relational rotation in complex
  space.
\newblock In \emph{7th International Conference on Learning Representations,
  {ICLR} 2019}.

\bibitem[{Vashishth et~al.(2020)Vashishth, Sanyal, Nitin, and
  Talukdar}]{Vashishth2020Composition-based}
Shikhar Vashishth, Soumya Sanyal, Vikram Nitin, and Partha Talukdar. 2020.
\newblock Composition-based multi-relational graph convolutional networks.
\newblock In \emph{International Conference on Learning Representations}.

\bibitem[{Vaswani et~al.(2017)Vaswani, Shazeer, Parmar, Uszkoreit, Jones,
  Gomez, Kaiser, and Polosukhin}]{DBLP:conf/nips/VaswaniSPUJGKP17}
Ashish Vaswani, Noam Shazeer, Niki Parmar, Jakob Uszkoreit, Llion Jones,
  Aidan~N. Gomez, Lukasz Kaiser, and Illia Polosukhin. 2017.
\newblock Attention is all you need.
\newblock In \emph{Advances in Neural Information Processing Systems {NIPS}
  2017}, pages 5998--6008.

\bibitem[{Velickovic et~al.(2018)Velickovic, Cucurull, Casanova, Romero,
  Li{\`{o}}, and Bengio}]{DBLP:conf/iclr/VelickovicCCRLB18}
Petar Velickovic, Guillem Cucurull, Arantxa Casanova, Adriana Romero, Pietro
  Li{\`{o}}, and Yoshua Bengio. 2018.
\newblock Graph attention networks.
\newblock In \emph{6th International Conference on Learning Representations,
  {ICLR}}.

\bibitem[{Vrandecic and Kr{\"{o}}tzsch(2014)}]{DBLP:journals/cacm/VrandecicK14}
Denny Vrandecic and Markus Kr{\"{o}}tzsch. 2014.
\newblock Wikidata: a free collaborative knowledgebase.
\newblock \emph{Commun. {ACM}}, 57(10):78--85.

\bibitem[{Wang et~al.(2021)Wang, Wang, Lyu, and Zhu}]{GRAN}
Quan Wang, Haifeng Wang, Yajuan Lyu, and Yong Zhu. 2021.
\newblock \href {https://doi.org/10.18653/v1/2021.findings-acl.35} {Link
  prediction on n-ary relational facts: A graph-based approach}.
\newblock In \emph{Findings of the Association for Computational Linguistics:
  ACL-IJCNLP 2021}, pages 396--407, Online. Association for Computational
  Linguistics.

\bibitem[{Wen et~al.(2016)Wen, Li, Mao, Chen, and
  Zhang}]{DBLP:conf/ijcai/WenLMCZ16}
Jianfeng Wen, Jianxin Li, Yongyi Mao, Shini Chen, and Richong Zhang. 2016.
\newblock On the representation and embedding of knowledge bases beyond binary
  relations.
\newblock In \emph{Proceedings of the Twenty-Fifth International Joint
  Conference on Artificial Intelligence, {IJCAI} 2016}, pages 1300--1307.

\bibitem[{Yan et~al.(2022)Yan, Zhang, Sun, Xu, Jin, and Li}]{Hyper2}
Shiyao Yan, Zequn Zhang, Xian Sun, Guangluan Xu, Li~Jin, and Shuchao Li. 2022.
\newblock \href {https://doi.org/https://doi.org/10.1016/j.neucom.2022.04.026}
  {Hyper2: Hyperbolic embedding for hyper-relational link prediction}.
\newblock \emph{Neurocomputing}, 492:440--451.

\bibitem[{Yu and Yang(2021)}]{HyTransformer}
Donghan Yu and Yiming Yang. 2021.
\newblock \href {http://arxiv.org/abs/2104.08167} {Improving hyper-relational
  knowledge graph completion}.
\newblock \emph{CoRR}, abs/2104.08167.

\bibitem[{Zhang et~al.(2018)Zhang, Li, Mei, and Mao}]{DBLP:conf/www/ZhangLMM18}
Richong Zhang, Junpeng Li, Jiajie Mei, and Yongyi Mao. 2018.
\newblock Scalable instance reconstruction in knowledge bases via relatedness
  affiliated embedding.
\newblock In \emph{The World Wide Web Conference, {WWW} 2018}, pages
  1185--1194.

\end{thebibliography}

	\appendix
	\newpage
	\section*{Appendix} 
	
	\textbf{Implementation details}:
	We implement all experiments at dim size 200 for all model variants. We employ Adam~\citep{DBLP:journals/corr/KingmaB14} optimizer with learning rate at 0.0001. For datasets with a higher proportion of hyper-relational facts, we utilize more training epochs. We conduct 500 training epochs for WD50K\_100 and WD50K\_66. For the remaining datasets, we use 400 training epochs. The message passing infrastructure is implemented with PyTorch Geometric~\citep{DBLP:journals/corr/abs-1903-02428}. All training was conducted on a Tesla V100. all model instance with parameter size smaller than 12.9M. the longest training time is about 5 days on WikiPeople.
 
 The detail hyper-parameters setting are as following table:
	\begin{table}[!htbp]
		\centering
		\resizebox{\linewidth}{!}{
			\begin{tabular}{cc}
				\toprule 
				\textbf{hyper-parameter}&\textbf{value set}\\ \hline
				embedding dim & \{\textbf{200},150,100\} \\ \hline
				batch size & \{\textbf{128},256,\} \\ \hline
				ReSaE dropout & \{0.1,0.2,\textbf{0.3}\} \\ \hline
				$\Phi_q$ &  \textbf{mean} sum max \\ \hline
				$\gamma_r$ & \textbf{mean} sum concatenation \\ \hline
				$\gamma_v$ & \textbf{mean} sum concatenation \\ \hline
				$\Phi_{q2}$ & \textbf{mean} sum concatenation \\ \hline
				$act$  & \{elu, \textbf{tanh},sigmoid,relu,gelu\} \\ \hline
				$act_r$ & \{\textbf{not apply},elu,tanh,sigmoid,relu,gelu\} \\ \hline
				decoder layer num & \{1,\textbf{2},3\} \\ \hline
				alpha and beta initial & \{\textbf{(0.8,0.2)},(0.7,0.3),(0.6,0.4),(0.5,0.5)\} \\ \hline
				decoder typewise-pooling & \textbf{mean} max sum \\ \hline
				decoder $\phi$ & \textbf{concatenation} mean \\ \hline
				decoder hidden dim & \{128,256,\textbf{512}\} \\ \hline
				decoder  head num & \{2,\textbf{4}\} \\ \hline
				transformer dropout & \{\textbf{0.1},0.2,0.3\} \\ \hline
				label smoothing & \textbf{0.1} 0.0 \\ \hline
		\end{tabular}}
		\caption{\label{table4}
			Main hyperparameter search of our approach.The selected values are in \textbf{bold}.
		}
	\end{table}

        \textbf{Limitations}:
        ReSaE calculates attention score matrix for all relations set at once, it may encounter memory issues when encoding KGs containing a substantial number of relations. In the future, we will consider methods that can mitigate this issue.

    \begin{table}[!htbp]
\small
    \centering
    \resizebox{\linewidth}{!}{
    \begin{tabular}{rrrrrrrr}
    \toprule
    \multicolumn{1}{c}{Dataset} & \multicolumn{1}{c}{Num Facts} & \multicolumn{1}{c}{H-Facts with Q(\%)} & \multicolumn{1}{c}{Entities} & \multicolumn{1}{c}{Relations} & \multicolumn{1}{c}{Train} & \multicolumn{1}{c}{Valid} & \multicolumn{1}{c}{Test} \\
    \midrule
    \multicolumn{1}{c}{JF17K} & \multicolumn{1}{c}{100,947} & \multicolumn{1}{c}{46,320(45.9\%)} & \multicolumn{1}{c}{28,645} & \multicolumn{1}{c}{501} & \multicolumn{1}{c}{76,379} & \multicolumn{1}{c}{-} & \multicolumn{1}{c}{24,568} \\
    \multicolumn{1}{c}{WikiPeople} & \multicolumn{1}{c}{369,866} & \multicolumn{1}{c}{9,482(2.6\%)} & \multicolumn{1}{c}{34,839} & \multicolumn{1}{c}{178} & \multicolumn{1}{c}{294,439} & \multicolumn{1}{c}{37,715} & \multicolumn{1}{c}{37,712} \\
    \multicolumn{1}{c}{WD50K} & \multicolumn{1}{c}{236,507} & \multicolumn{1}{c}{32,167(13.6\%)} & \multicolumn{1}{c}{47,156} & \multicolumn{1}{c}{532} & \multicolumn{1}{c}{166,435} & \multicolumn{1}{c}{23,913} & \multicolumn{1}{c}{46,159} \\
    \bottomrule
    \end{tabular}
}
\caption{\label{table5}
The statistics of the main dataset, with columns indicating the number of facts, facts with qualifiers, entities, relations, and train/valid/test sets, respectively.}
\end{table}

\end{document}